# NeuroLingua: A Language-Inspired Hierarchical Framework for Multimodal Sleep Stage Classification Using EEG and EOG

Mahdi Samaee[2], Mehran Yazdi[*,1,2] and Daniel Massicotte[1]

[1]Laboratory of Signal and System Integration, Department of Electrical and Computer Engineering,
Université du Québec à Trois-Rivières, Trois-Rivières, Canada
[2]Signal and Image Processing Laboratory, School of Electrical and Computer Engineering, Shiraz University, Shiraz, Iran

mahdisamaee@hafez.shirazu.ac.ir, Mehran.Yazdi@uqtr.ca and Daniel.Massicotte@uqtr.ca

*Corresponding author

**Abstract**

Automated sleep stage classification from polysomnography remains limited by the lack of expressive temporal hierarchies, challenges in multimodal EEG and EOG fusion, and the limited interpretability of deep learning models. We propose NeuroLingua, a language-inspired framework that conceptualizes sleep as a structured physiological language. Each 30-second epoch is decomposed into overlapping 3-second subwindows ("tokens") using a CNN-based tokenizer, enabling hierarchical temporal modeling through dual-level Transformers: intra-segment encoding of local dependencies and inter-segment integration across seven consecutive epochs (3.5 minutes) for extended context. Modality-specific embeddings from EEG and EOG channels are fused via a Graph Convolutional Network, facilitating robust multimodal integration. NeuroLingua is evaluated on the Sleep-EDF Expanded and ISRUC-Sleep datasets, achieving state-of-the-art results on Sleep-EDF (85.3% accuracy, 0.800 macro F1, and 0.796 Cohen's κ), and competitive performance on ISRUC (81.9% accuracy, 0.802 macro F1, and 0.755 κ), matching or exceeding published baselines in overall and per-class metrics. The architecture's attention mechanisms enhance the detection of clinically relevant sleep microevents, providing a principled foundation for future interpretability, explainability and causal inference in sleep research. By framing sleep as a compositional language, NeuroLingua unifies hierarchical sequence modeling and multimodal fusion, advancing automated sleep staging toward more transparent and clinically meaningful applications.

## 1. Introduction

Sleep is a vital neurophysiological process fundamental to human health, cognitive function, and emotional regulation [10]. Accurate classification of sleep stages—encompassing Wake (W), Non-Rapid Eye Movement (NREM: stages N1, N2, N3), and Rapid Eye Movement (REM)—is essential for the diagnosis of sleep disorders, evaluation of treatment efficacy, and exploration of brain dynamics during rest [10, 16]. Standard practice relies on expert manual annotation of polysomnographic (PSG) recordings, which include multi-channel signals such as electroencephalogram (EEG), electrooculogram (EOG), and electromyogram (EMG), following well-defined clinical guidelines (e.g., those of the American Academy of Sleep Medicine [19]). However, manual scoring is time-consuming, subject to inter-rater variability, and not scalable to the growing demand for sleep analysis, underscoring the need for automated, reliable, and interpretable sleep stage classification systems. Such systems carry significant clinical and societal relevance, promising to alleviate the burden on sleep laboratories and enable broader access to sleep disorder diagnostics.

Recent advances in deep learning have indeed propelled the development of automated sleep staging algorithms that learn complex patterns directly from raw PSG data [22, 23]. These data-driven models have significantly improved classification performance compared to traditional feature-engineered methods. Nonetheless, critical gaps remain in current approaches. Many state-of-the-art models treat sleep data as a flat temporal sequence, overlooking the rich hierarchical structure of sleep architecture. In reality, sleep is an inherently multi-scale phenomenon: transient micro-events (e.g., sleep spindles, K-complexes, rapid eye movements) on the order of 0.5–2 seconds are embedded within broader oscillatory patterns that define each 30-second epoch and beyond.

This compositional structure of micro- and macro-events invites a conceptual analogy to natural language, where basic tokens (phonemes or words) combine to form higher-order syntax (phrases, sentences) characterizing a narrative. Equally important, most deep learning models for sleep staging operate as black boxes with limited transparency, lacking interpretability to explain their decisions – a liability in clinical settings where trust and insight are paramount [24]. These limitations in hierarchical temporal modeling, modality fusion, and model interpretability motivate a fundamentally new approach to automated sleep staging. Notably, the proposed framework's explicit hierarchical design and attention mechanisms inherently offer strong potential for future interpretability and explainability, paving the way for more transparent and clinically trustworthy sleep stage classification.To address the persistent challenges in automatic sleep stage classification—including limited interpretability, weak modeling of temporal hierarchies, and inefficient multimodal fusion—we propose **NeuroLingua**, a novel and theoretically grounded framework inspired by the structural organization of natural language. NeuroLingua reconceptualizes sleep analysis as a form of language processing, where the architecture of sleep is modeled analogously to the grammar of language: subwindows as tokens, 30 s epochs as sentences, and multi-epoch sequences as paragraphs. This linguistically motivated framing enables the model to capture both local signal dynamics and long-range temporal dependencies, which are often oversimplified in conventional approaches.

We develop and analyze **NeuroLingua**, our main contribution, which integrates three core innovations:

1. **A Convolutional Neural Networks (CNN)-based tokenization module** that extracts localized features from short subwindows of EEG/EOG signals (akin to syllables in a sleep language);

2. **A dual-level Transformer architecture** that models intra-epoch and inter-epoch dependencies (similar to syntactic and narrative parsing in language);

3. **A Graph Neural Network (GNN)-based cross-modal fusion strategy** that integrates EEG and EOG in a topology-aware manner, capturing inter-channel relationships beyond naïve concatenation.

This design enables rich, multi-level temporal representation learning and principled multimodal integration, while also providing potential interpretability and explainability through attention maps and graph connectivity patterns that align with physiological events.

We evaluate NeuroLingua on two publicly available datasets— **Sleep-EDF Expanded [24, 25]** and **ISRUC-Sleep [26]**—demonstrating robust and generalizable performance across both. Moreover, the hierarchical and modular nature of our framework paves the way for future directions, including interpretable and explainable AI and causal inference integration, positioning NeuroLingua as a flexible foundation for next-generation, clinically aligned sleep staging systems.

**Our key contributions are:**
- **Proposing NeuroLingua**, a linguistically inspired framework for sleep stage classification that reinterprets neurophysiological signal dynamics through the structural hierarchy of natural language—modeling subwindows as tokens, segments as sentences, and temporal sequences as paragraphs. This perspective enables richer contextual understanding and structured temporal representation.

- **Developing NeuroLingua**, a high-capacity hierarchical architecture that combines CNN-based tokenization, Transformer-based sequence modeling at both intra- and inter-segment levels, and Graph Neural Network-based modality fusion. This model achieves state-of-the-art performance and is designed to support future advancements in interpretability and causality-aware analysis.

- **Demonstrating robust performance across benchmark datasets** (Sleep-EDF Expanded and ISRUC-Sleep), with NeuroLingua achieving superior accuracy—especially in difficult-to-classify stages—thereby advancing the field of automatic sleep staging.

- **Establishing a strong foundation for future research**, as NeuroLingua provides an extensible platform for interpretability-driven and causality-based modeling, expanding the clinical and real-world impact of automated sleep staging systems.

2. Literature Review

The evolution of automatic sleep stage classification has transitioned from traditional feature-engineered paradigms to powerful deep learning architectures. Early works relied heavily on handcrafted features derived from EEG, EOG, and EMG signals—such as power spectral density, entropy, and statistical descriptors—and used classical classifiers including Support Vector Machines (SVMs), Random Forests, and Hidden Markov Models (HMMs) [23, 27]. Alickovic and Subasi [20] introduced an ensemble SVM framework to enhance robustness across subjects, while Ghasemzadeh et al. [28] applied a non-linear Logistic Smooth Transition Autoregressive (LSTAR) model for temporal dynamics. Al-Salman et al. [29] combined clustering-based probability distribution features with traditional classifiers for improved performance. However, these early approaches were limited by modest accuracy, poor generalization, and difficulty in distinguishing transitional stages such as N1.

The advent of deep learning brought a paradigm shift, enabling direct learning from raw or minimally preprocessed signals. CNNs quickly became dominant due to their ability to automatically capture local signal characteristics, such as spindles and K-complexes. Supratak et al. introduced DeepSleepNet [30], a hybrid CNN-LSTM (Long Short-Term Memory) model for raw EEG processing, followed by SeqSleepNet [31], which embedded a temporal hierarchy using attention-guided recurrent units. Dong et al. [32] proposed a mixed CNN-RNN (Recurrent Neural Network) architecture that modeled short- and long-term patterns jointly. Goshtasbi et al. [33] introduced SleepFCN, a fully convolutional model optimized for single-channel EEG, while Zhang and Wu [34] explored unsupervised complex-valued CNNs for spatial representation learning. Zhu et al. [35] enhanced CNN-based models with temporal attention to capture stage-specific dependencies more effectively.

Beyond single-channel approaches, multimodal and multivariate inputs have gained attention. Chambon et al. [36] developed a deep learning pipeline that exploited cross-channel temporal correlations in multivariate PSG data. Kim et al. [37] employed a multi-level fusion mechanism to combine EEG, EOG, and EMG representations, while Ji et al. [18] proposed 3DSleepNet, which used structured 3D modeling of multi-channel biosignals. These approaches highlight the importance of cross-modal interactions for robust classification.

Transfer learning and self-supervised techniques have also gained traction for improving generalization and reducing dependency on large labeled datasets. Guillot and Thorey [38] presented RobustSleepNet, which leverages transfer learning at scale. Radha et al. [39] demonstrated effective adaptation from EEG to (photoplethysmogram) PPG-based wearable sleep staging. Eldele et al. [40] offered a comprehensive evaluation of self-supervised learning strategies, showcasing the feasibility of label-efficient training in clinical settings.

RNNs, especially LSTM and (Gated Recurrent Unit) GRU-based models, have been widely employed for their temporal modeling capabilities. However, these models are inherently sequential, which limits their computational efficiency and hinders their ability to handle long-range dependencies. Li et al. [14] addressed these issues using an LSTM-Ladder architecture to enhance hierarchical representation. Moctezuma et al. [41] proposed a GRU-powered model with permutation-based EEG channel selection to reduce redundancy and improve generalization. Huang et al. [42] integrated a transition-optimized Hidden Markov Model (HMM) to smooth predictions and more accurately capture sleep stage transitions.

Transformer-based architectures have emerged as state-of-the-art due to their self-attention mechanism, which effectively captures both short- and long-range dependencies in parallel. Phan et al. introduced SleepTransformer [21], offering interpretability and uncertainty quantification. Pradeepkumar et al. [43] developed a cross-modal Transformer that leveraged multiple signal types, while Lee et al. [44] proposed SleepXViT, incorporating explainability and spatial structure into a vision-based Transformer. Dai et al. [45] presented MultiChannelSleepNet, a Transformer-based model tailored for full PSG inputs. Dutt et al. [46] advanced explainability through attention heatmaps in SleepXAI, while Jha et al. [47] introduced SlumberNet, a residual network-based Transformer variant for robust stage detection. Cho and Lee [48] combined residual blocks with directed transfer functions to enrich EEG connectivity modeling. Despite their accuracy, Transformers are computationally expensive; Wenjian et al. [49] addressed this with DynamicSleepNet, a multi-exit design that adapts inference depth based on sample complexity.

Temporal convolutional models offer an efficient alternative to recurrent and Transformer-based architectures. Khalili and Asl [50] utilized Temporal Convolutional Networks (TCNs) with innovative data augmentation to achieve competitive performance on single-channel EEG. Their design improved training stability and reduced overfitting.

Concurrently, graph-based methods have been explored for spatially structured sleep modeling. Jia et al. introduced GraphSleepNet [51], which dynamically learned inter-channel dependencies via spatial-temporal graph convolutions. Wang et al. [52] extended this by employing multi-layer Graph Attention Networks (GATs) to prioritize salient brain regions during classification. These models enhance robustness to subject variability and naturally support multimodal fusion.

Despite these developments, many prior approaches treat sleep as a flat sequence, ignoring its rich temporal hierarchy. In reality, sleep architecture comprises nested events—microphenomena, such as spindles and K-complexes, embedded within macro-stage dynamics. To address this, we propose **NeuroLingua**, a linguistically inspired framework that draws structural parallels between language and sleep. It uses CNN-based tokenization to extract subwindow-level representations, dual-level Transformers to model intra- and inter-segment dependencies, and GNN-based fusion for modality integration. This multi-scale modeling approach enables the capture of both fine-grained transitions and long-range contextual patterns, thereby improving classification accuracy and interpretability across datasets such as Sleep-EDF and ISRUC.

**3. NeuroLingua Method**

This section details the architecture and implementation of NeuroLingua. Rooted in a linguistic analogy to sleep architecture, NeuroLingua is a hierarchically structured deep learning model designed for automated classification of sleep stages. Drawing inspiration from the structured nature of natural language and the biological dynamics of sleep, the model systematically transforms raw electrophysiological inputs into increasingly abstract, temporally contextualized, and semantically meaningful representations.

The framework integrates several key components:
(i) A standardized preprocessing pipeline;
(ii) CNN-based tokenization to extract localized features from physiological subwindows;
(iii) Dual-level Transformer encoders for capturing both intra- and inter-segment dependencies; and
(iv) Graph-based multimodal fusion using a Graph Convolutional Network (GCN) to integrate EEG and EOG modalities in a topology-aware fashion.

Each stage is carefully designed to maintain the physiological fidelity of the input signals while enabling deep temporal modeling and inter-modal reasoning. Figure 1 provides an overview of the proposed method. The following subsections elaborate on each component of this architecture.

To clarify our notation, we denote the $i-th$ 30 s sleep segment for each signal modality (e.g., EEG1, EEG2, EOG) as $X_i^{(m)} \in \mathbb{R}^{1 \times T}$, where $m$ indexes the modality and $T$ is the number of time samples (3,000 at 100 Hz). Each segment is divided into overlapping subwindows, with the $j-th$ subwindow for modality $m$ represented as $x_j^{(m)}$. The CNN outputs an embedding $z_j^{(m)}$ for each subwindow, which is then processed by the intra-segment Transformer to yield $f_i^{(m)}$. These representations are further modeled by the inter-segment Transformer, resulting in high-level, context-aware embeddings $h_i^{(m)}$. For clarity and to avoid notational clutter, the modality index $m$ is omitted in subsequent formulas, though all variables are in fact defined separately for each modality prior to multimodal fusion.

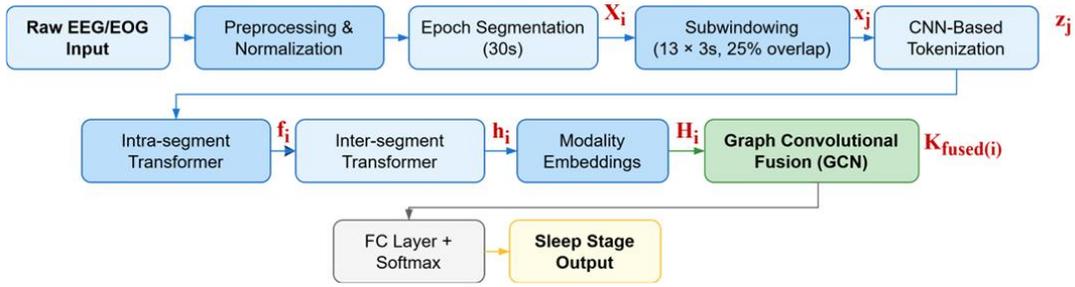

*Figure 1. Overview of the proposed method. Each 30 s segment ($X_i$) is divided into overlapping 3 s subwindows($x_j$), tokenized by a CNN ($z_j$), and processed through intra-($f_i$) and inter-segment($h_i$) Transformers. Modality embeddings are fused via GCN to produce the final representation ($K_{fused(i)}$).*

### 3.1. Preprocessing

All input signals, comprising two EEG channels as well as EOG, undergo a uniform preprocessing pipeline designed to mitigate artifacts and preserve physiologically relevant information. Each modality is processed independently. Initially, signals are bandpass filtered with a zero-phase, fifth-order Butterworth filter (0.5–49.9 Hz; 100 Hz sampling rate) to remove baseline drift and high-frequency noise while retaining critical spectral components such as delta, spindle, and alpha rhythms. The filtered signals are subsequently partitioned into consecutive, non-overlapping 30 s epochs, following the American Academy of Sleep Medicine (AASM) [3] guidelines. These epochs serve as the fundamental modeling units, providing the temporal granularity for both expert annotation and automated classification.

### 3.2. Language-Inspired Hierarchical Temporal Decomposition for EEG/EOG

Inspired by Natural Language Processing (NLP), we introduce a hierarchical temporal segmentation strategy within our NeuroLingua framework for EEG/EOG-based sleep stage classification. NeuroLingua draws a structural analogy between linguistic constructs and physiological time series, wherein *paragraphs*, *sentences*, and *tokens* correspond to progressively finer-grained temporal units in EEG/EOG signals. At the highest level of this hierarchy, continuous EEG/EOG recordings are partitioned into non-overlapping 30 s epochs. Each epoch is treated as a *sentence* ($X_i$), aligning with conventional sleep scoring intervals. Successive sequences of these 30 s epochs are grouped into larger temporal contexts, referred to as *paragraphs*, which enable the modeling of long-range inter-epoch dependencies—akin to how paragraphs organize coherent thoughts in language. Within each 30 s epoch, we perform a finer decomposition into thirteen overlapping 3 s subwindows—termed *tokens* ($x_j$). Each token comprises 300 samples and is intended to capture short-duration physiological micro-events such as spindles, K-complexes, and REM bursts. This decomposition forms the basis of intra-segment modeling. The parameters for tokenization were determined empirically. We evaluated multiple subwindow lengths and overlap configurations, ultimately selecting a 3 s window with 25% overlap (stride of 2.25 s). As shown in Table 1, we further analyze and discuss why this specific duration is particularly effective, demonstrating that it provides a favorable trade-off between temporal resolution and the ability to detect sleep-relevant microstructures. Formally, for a 30 s signal segment $X$, we extract 13 overlapping subwindows $x_i$ using:

$$x_j = X(t_j : t_j + W), \quad (1)$$

where $t_j = j \times S$, $j = 0, 1, \dots, 12$, and with time windows $W = 3\ s$ and stride $S = 2.25\ s$. This results in a sequence of 13 temporal "tokens" per segment. This hierarchical segmentation process is visually illustrated in Figure 2. Figure 3 visually summarizes this multi-scale decomposition, aligning EEG/EOG segmentation with linguistic structure: continuous signals are divided into sentences ($X_i$), each of which is decomposed into overlapping tokens ($x_j$) (subwindows), and grouped into paragraphs (temporal contexts). This hierarchical design allows NeuroLingua to effectively model both local (intra-segment) and global (inter-segment) temporal patterns, enhancing classification performance by leveraging the full spectrum of dynamics inherent in sleep physiology.

### 3.3. Multi-Scale Convolutional Feature Extraction

Following the hierarchical segmentation described in the previous section, each 3 s subwindow (token, $x_j$) undergoes modality-specific convolutional encoding to transform raw physiological data into compact, semantically rich representations. Separate encoders are designed for each input modality (e.g., EEG and EOG), enabling the extraction of modality-tailored features from each subwindow. This step functions as a neural tokenization mechanism: each raw token is converted into a fixed-dimensional feature embedding that encapsulates local temporal patterns relevant for sleep staging. These embeddings constitute the model's learned vocabulary of neural tokens and form the foundational input for the subsequent temporal modeling modules (see Section 3.4.).

#### 3.3.1. Modality-Specific Feature Extractors

Each 3 s subwindow from each modality is independently transformed into a fixed-dimensional embedding via a dedicated multi-scale CNN. The CNN comprises four parallel convolutional branches with kernel sizes of 16, 32, 64, and 128. The selection of kernel sizes is physiologically motivated and aligns with the typical durations of canonical sleep microstructures. As detailed in Table 2, each kernel size targets specific neurophysiological events—ranging from brief vertex waves and spindle onsets to slower phenomena such as delta waves and slow eye movements—ensuring comprehensive coverage of temporal dynamics relevant to sleep staging. The outputs of these branches are concatenated and passed through a sequence of additional operations, including convolution, batch normalization, ReLU activation, and max-pooling. Finally, an adaptive average pooling layer compresses the resulting feature maps into a unified embedding vector for each subwindow. This multi-scale architecture is theoretically motivated to capture both short and long-duration sleep micro-events, such as spindles, K-complexes, and REM bursts. Empirical evaluation confirms that this design consistently outperforms single-scale alternatives in classification accuracy. Formally, the CNN encoder maps each subwindow $x_j$ to a feature token $z_j \in \mathbb{R}^{d_{cnn}}$, where $d_{cnn}$ denotes the CNN based feature embedding dimension ($d_{cnn} = 64$). This dimensionality was chosen empirically based on a trade-off between information preservation and computational efficiency. It is sufficient to capture local spectral and temporal dynamics, while maintaining a manageable sequence length and feature size for the subsequent Transformer layers. The resulting sequence $\{z_0, z_1, ..., z_{12}\}$ per segment serves as input to the temporal modeling pipeline. The structure of the multi-scale CNN encoder is depicted in Figure 4.

Table 1. Sleep Microstructures and Their Capturability in 3-Second Subwindows.

| Micro-structure | Ref. | Description | Typical Duration | Capturable in 3 s Window? | Clinical Relevance |
|---|---|---|---|---|---|
| Sleep spindle | [3, 4] | 11–16 Hz rhythmic burst, waxing-waning amplitude | ~0.5–2 s | Yes | Marker of N2 stage; linked to memory consolidation |
| K-complex | [3, 7] | High-amplitude biphasic waveform | ~0.5–1 s | Yes | Common in N2; thought to suppress arousals |
| REM bursts | [3, 13] | Rapid eye movement episodes (EOG) | 0.3–2 s per burst | Yes | Indicative of REM sleep |
| Vertex sharp wave | [3] | Sharp transient, maximal at vertex (Cz) | ~0.2–0.5 s | Yes | May occur in N1 and early N2 |
| Arousal | [3] | Brief return to lighter stage or wake | <15 s (often <5 s) | Partial | Reflects sleep fragmentation or instability |
| Delta burst | [3] | High-amplitude, low-frequency (<4 Hz) activity | ~1–2 s per burst | Yes | Common in N3; defines slow-wave sleep |
| Eye blinks (wake) | [3] | Slow upward/downward deflections in EOG | ~0.5–1 s | Yes | Common during wake or drowsy states |

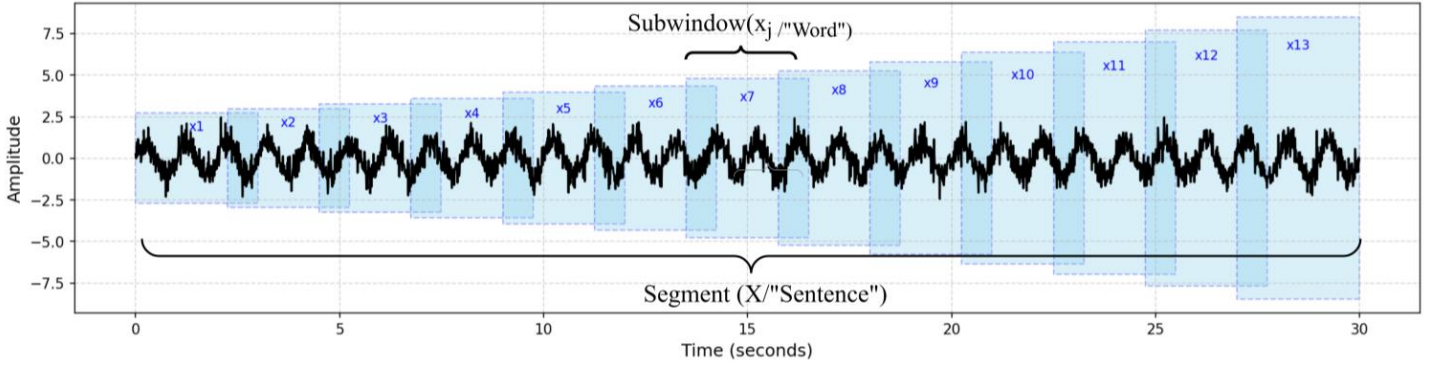

Figure 2. Segmentation of 30 s Signal (X/"Sentence") into 3s Windows (xj/"Word") with 25% Overlap.

### 3.4. Hierarchical Temporal Modeling with Transformers

To capture the rich multi-scale temporal dynamics present in sleep signals, NeuroLingua employs a two-level Transformer-based architecture that explicitly models both short-term (intra-segment) and long-term (inter-segment) dependencies. This design is grounded in the physiological and clinical understanding that sleep stage classification inherently depends on temporal context at multiple scales, as highlighted in the American Academy of Sleep Medicine (AASM) guidelines [3]. According to the AASM, accurate annotation of sleep stages requires consideration of not only local signal features within a 30 s epoch but also patterns and transitions spanning several contiguous epochs, since certain stages (e.g., N2, REM) are characterized by specific microstructures and sequential evolution over time [3, 53]. Intra-segment modeling focuses on extracting short-term dependencies and capturing transient sleep micro-events (such as spindles, K-complexes, or rapid eye movement bursts) that are often confined within a single epoch or subwindow [54]. These microstructures are critical for stage identification, particularly in distinguishing between light and deep sleep or detecting REM periods [3, 54]. In contrast, inter-segment modeling captures longer-term contextual dependencies by integrating information across multiple consecutive epochs, which is essential for accurately detecting stage transitions and for disambiguating epochs that may exhibit ambiguous or borderline features in isolation [3, 55]. Both levels in our architecture utilize the same core Transformer encoder structure, consisting of two stacked layers, each with multi-head self-attention, feedforward sublayers, residual connections, and layer normalization, as proposed by Vaswani et al. [56]. Each Transformer module receives a sequence of modality-specific input embeddings—subwindow-level for intra-segment modeling and segment-level for inter-segment modeling—augmented with learnable positional encodings to preserve temporal order. The outputs are subsequently processed via adaptive pooling to yield fixed-dimensional, context-aware representations. This unified Transformer design enables the model to jointly learn localized patterns (within a 30 s segment) and broader transitions across multiple segments (reflecting stage continuity and sleep architecture), thereby enhancing its ability to classify sleep stages with both precision and contextual awareness. Figure 5 depicts the structure of the proposed Transformer-based network.

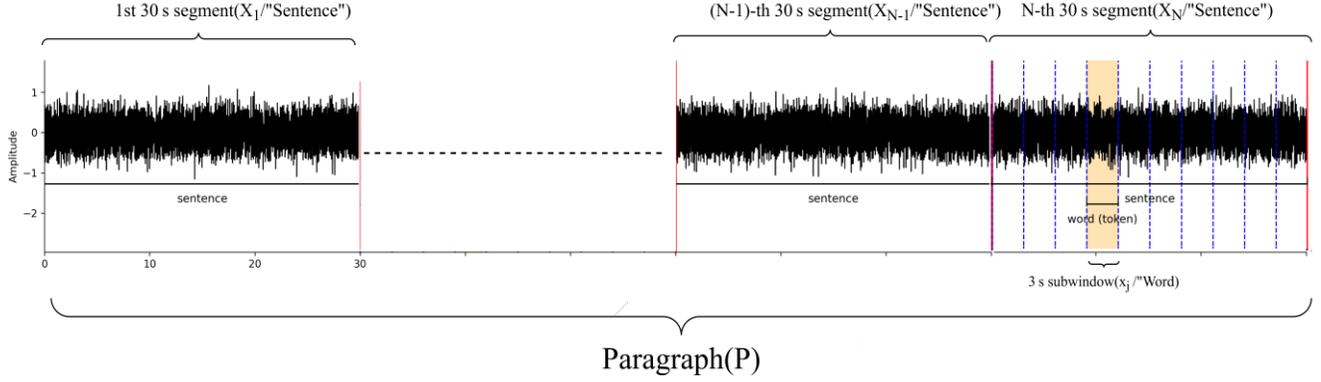

Figure 3. Hierarchical Language-Inspired Representation of sleep Signals: each 3 s subwindow($x_j$/ "$Word$")forms part of a 30 s segment ($X_i$/ "Sentence"); consecutive segments form a "paragraph" ($P$).

Table 2. Multi-Scale Convolutional Kernel Sizes and Their Physiological Relevance

| Kernel Size (Samples) | Approx. Duration @ 100 Hz | Captured Microstructures | Scientific Justification |
|---|---|---|---|
| 16 | 160 ms | Vertex sharp waves, eye blinks | Enables detection of very short events like vertex sharp waves, which occur at subsecond scales. Provides fine temporal resolution. |
| 32 | 320 ms | Spindle onsets/offsets, small phasic events | Captures transition points of spindles, onset of K-complexes, and minor phasic events. Useful for precise timing analysis in both EEG and EOG. |
| 64 | 640 ms | Sleep spindles, K-complexes, rapid eye movements (REMs) | Covers the full duration of canonical spindles (~0.5–2 sec) and K-complexes (~0.5–1 sec), critical for N2 stage. Also suitable for detecting REM bursts. |
| 128 | 1.28 sec | Delta waves, slow eye movements (SEMs), full spindles | Enables the detection of slower, extended events like delta waves (<4 Hz) and slow eye movements, essential for N1 and N3 stage recognition. |

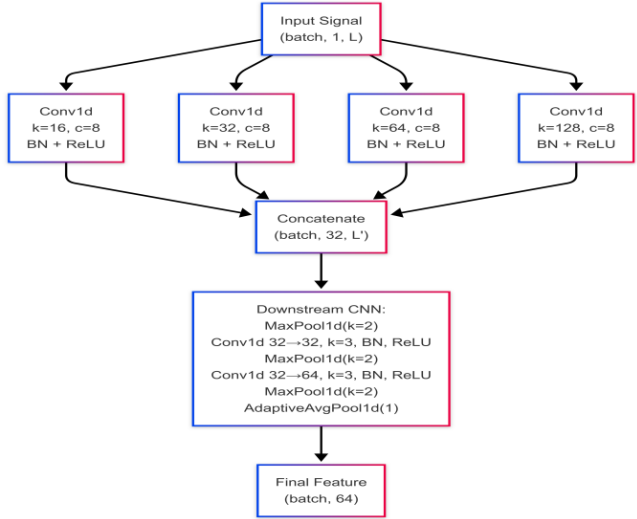

Figure 4. proposed multi kernel CNN.

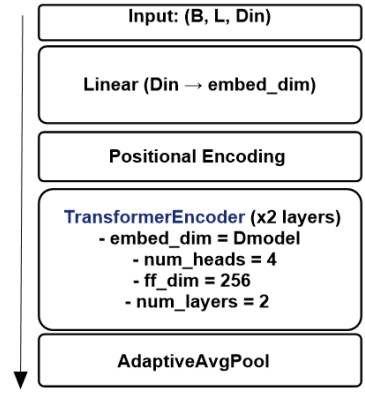

Figure 5. proposed transformer based network for intra and inter temporal modeling.

### 3.4.1. Intra-Segment Temporal Modeling

For each 30 s segment, a modality-specific Transformer processes the sequence of 13 subwindow embeddings (tokens), generated as described in Section 3.3. These embeddings are first linearly projected into a shared latent space and augmented with positional encodings. The Transformer encoder models dependencies between subwindows—such as a spindle following a K-complex—and outputs a contextualized sequence, which is then aggregated using adaptive average pooling to yield a fixed-length embedding representing the entire segment.

Formally, let

$$Z_i = \{z_0, z_1, \ldots, z_{12}\} \in \mathbb{R}^{13 \times d_{cnn}} \quad (2)$$

be the subwindow embeddings for the $i - th$ segment of a given modality ($d_{cnn} = 64$). The Transformer encoder $T(\cdot)$ produces:

The input to the Transformer module is a sequence of feature embeddings with shape $(B, L, D_{in})$, where $B$ is the batch size, $L$ is is the sequence length and $D_{in} = 64$ is the feature dimension output by the preceding CNN. This input first passes through a linear layer that projects each embedding from dimension $D_{in} = 64$ to the Transformer embedding dimension, which is set to 128 in our implementation. It is a common practice to set the Transformer embedding dimension equal to or greater than the input feature dimension [56, 57]. This allows the model to project lower-dimensional features into a richer embedding space before sequential modeling. Such expansion enhances the model's ability to capture complex temporal relationships while maintaining efficient computation and facilitating residual connections.

$$f_i = T_{intra}(Z_i) \qquad (3)$$

where $f_i \in \mathbb{R}^{d_{tr}}$ is the resulting segment-level embedding ($d_{tr}$=128).

### 3.4.2. Inter-Segment Temporal Modeling

To model longer-range transitions and contextual patterns across multiple epochs, NeuroLingua applies a second Transformer encoder to a sequence of seven consecutive segment embeddings, representing a 3.5-minute context window (which is considered as "paragraph" here). Each modality is modeled independently, and positional encodings are again used to retain segment order. The choice of using a window of seven segments was empirically determined. We systematically evaluated multiple window sizes and found that modeling seven consecutive segments consistently yielded the best trade-off between performance and computational efficiency, providing sufficient temporal context for capturing sleep stage transitions such as N2 to REM or REM to N1. This grouping of seven consecutive segments is visually illustrated in Figure 6, which demonstrates how the model constructs a 3.5-minute context window for inter-segment temporal modeling. Let $f_i \in \mathbb{R}^{d_{tr}}$ be the segment embedding for the $i$-th segment of modality $m$. The input sequence:

$$F_i = [f_{i-6}, ..., f_{i-1}, f_i] \in \mathbb{R}^{7 \times d_{tr}} \qquad (4)$$

is processed to produce a global, context-enriched representation of the 3.5-minute sleep window (paragraph) as follows:

$$h_i = T_{inter}(F_i) \qquad (5)$$

where $h_i \in \mathbb{R}^{d_{tr}}$ is the resulting paragraph-level embedding ($d_{tr}$=128).

### 3.5. Adaptive Multimodal Fusion via Graph Convolutional Networks

Following the hierarchical temporal modeling described in Section 3.4, NeuroLingua performs modality-level fusion to integrate contextualized information from the two EEG channels and the EOG channel. To achieve this, we adopt a graph-based strategy that models inter-modality relationships as message-passing operations over a fully connected graph, allowing the network to dynamically adjust the contribution of each modality based on physiological context.

**Modality Embedding Graph Construction:** Each modality produces a global context embedding through its respective inter-segment Transformer encoder. These three embeddings are stacked to form a modality embedding matrix:

$$H_i = \begin{bmatrix} h_i^{EEG1} \\ h_i^{EEG2} \\ h_i^{EOG} \end{bmatrix} \in \mathbb{R}^{3 \times d_{tr}} \qquad (6)$$

where $h_i^{EEG1}, h_i^{EEG2}, h_i^{EOG} \in \mathbb{R}^{d_{tr}}$ are the segment-level embeddings for the $i$-th segment from the inter-segment Transformer and $d_{tr} = 128$ is the embedding dimension. To model the interactions between modalities, we define a fully connected, undirected graph $G = (V, \varepsilon)$ where each node $v_i \in V$ corresponds to one modality, and $\varepsilon = \{(a,b) | a,b \in \{1,2,3\}\}$ is the set of edges connecting every pair of modalities. This results in a complete graph of three nodes. The input to the graph convolutional network is the pair $(H_i, \varepsilon)$, where, $H_i \in \mathbb{R}^{3 \times 128}$ is the node feature matrix (modality embeddings) and $\varepsilon$ is the edge index tensor representing all possible pairs in the fully connected graph.

**Graph Convolutional Fusion:** We implement a two-layer GCN based on the formulation by Kipf and Welling [58], designed to propagate information across modality nodes and learn a fused representation. The network architecture is as follows:

- **Layer 1**: GCNConv(128 → 128), followed by ReLU activation
- **Layer 2**: GCNConv(128 → 128), producing updated node embeddings

Mathematically, the fusion is defined as:

$$K_i^{(1)} = \text{ReLU}(\text{GCN}_1(H_i, \varepsilon)), \quad K_i^{(2)} = \text{GCN}_2(K_i^{(1)}, \varepsilon) \qquad (7)$$

where $K^{(2)} \in \mathbb{R}^{3 \times 128}$ contains the fused, context-enriched embeddings for each modality after graph message passing. To obtain a unified representation for classification, the final embedding is computed as the mean across all modality nodes:

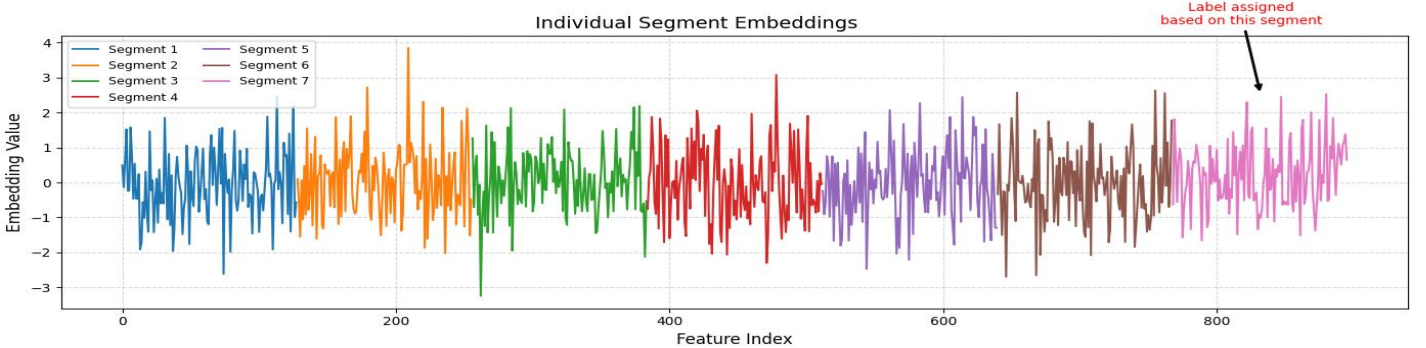

*Figure 6. Visual Representation of Segment Embeddings for a 3.5-Minute Context Window.*

$$K_{\text{fused}(i)} = \frac{1}{3}\sum_{m=1}^{3} k_i^{(2)}(m) \in \mathbb{R}^{128} \quad (8)$$

This fused embedding aggregates both intra- and inter-modality temporal context, capturing complementary information from EEG and EOG signals and enabling robust downstream classification.

### 3.6. Classification and Training Protocol

The fused multimodal embedding is forwarded to a fully connected layer, followed by a softmax classifier to predict the probability distribution over five sleep stages: Wake, N1, N2, N3, and REM. The entire NeuroLingua framework is trained end-to-end using the cross-entropy loss function, which is a standard choice for multi-class classification tasks and encourages the model to assign high probability to the correct sleep stage for each input segment. Model parameters—including those of the CNNs, Transformers, and GCN—are updated jointly via backpropagation and optimized using the Adam optimizer with a learning rate of 0.001 and a weight decay of $10^{-4}$. Adam is selected for its ability to adapt learning rates for individual parameters, resulting in faster convergence and improved performance on deep networks. The weight decay term acts as an L2 regularizer, helping to mitigate overfitting by penalizing large weights. To further enhance training stability and performance, a step-based learning rate scheduler is employed. This scheduler reduces the learning rate by a factor of 0.1 every 5 epochs, allowing the model to make larger updates in the early stages of training and more refined adjustments as it approaches convergence. This strategy helps prevent the optimizer from getting stuck in local minima and supports better generalization. To ensure robust generalization and avoid subject-specific overfitting, a rigorous subject-wise cross-validation protocol is employed in all experiments.

**Summary** – By unifying multi-scale CNN-based tokenization, hierarchical temporal modeling through dual-level Transformers, and adaptive multimodal fusion via graph neural networks, NeuroLingua provides a comprehensive and extensible solution for sleep stage classification. Its design is informed by both physiological insights and computational principles, enabling robust performance, interpretability, explainability, and seamless integration of domain knowledge.

## 4. Experimental Results

This section presents a comprehensive evaluation of the proposed NeuroLingua framework on two widely recognized benchmark datasets for sleep stage classification: Sleep-EDF Expanded and ISRUC-Sleep Subset 1. We detail our data selection, preprocessing, and rigorous subject-wise cross-validation strategy, followed by a thorough analysis of classification results, including both quantitative metrics and qualitative sequence-level comparisons. Our aim is to assess the robustness, generalizability, and practical applicability of NeuroLingua across healthy and clinical populations.

### 4.1. Datasets and Evaluation Strategy

NeuroLingua is evaluated on two benchmark datasets: Sleep-EDF Expanded [24, 25] and ISRUC-Sleep Subset 1 [26]. The Sleep-EDF Expanded dataset includes the "Sleep Cassette" subset, from which two EEG channels and one EOG channel, all sampled at 100 Hz, are selected for analysis. ISRUC-Sleep Subset 1 comprises recordings from 100 clinical subjects, with two EEG channels and one EOG channel selected for analysis. These channels were originally sampled at 200 Hz and subsequently downsampled to 100 Hz for consistency. The distributions of sleep stages across both datasets are presented in Table 3. To ensure subject-independent evaluation and assess the model's ability to generalize across unseen individuals, we adopted a rigorous 20-fold cross-validation strategy with a subject-wise split. This approach guarantees that data from a single subject is never shared between the training and validation folds.

*Table 3. Distribution of 30-second Sleep Stage Samples Across Datasets*

| Dataset | Wake | N1 | N2 | N3 | REM | Total |
|---|---|---|---|---|---|---|
| EDF-X | 65795 | 21469 | 68633 | 12991 | 25767 | 194655 |
| ISRUC-S1 | 20098 | 11062 | 27511 | 17251 | 11265 | 87187 |

The classification performance is reported in terms of overall accuracy (ACC), macro-averaged F1-score (MF1), Cohen's kappa coefficient (κ), and per-class F1 for each sleep stage (Wake, N1, N2, N3, REM). For completeness, the performance metrics are defined as follows. Let $TP_i$, $FP_i$, $TN_i$ and $FN_i$ denote, respectively, the number of true positives, false positives, true negatives, and false negatives for class $i$, as follows

$$ACC = \frac{\sum_{i=1}^{K} TP_i}{M}, \quad MF1 = \frac{1}{K}\sum_{i=1}^{K} \frac{2*Precision_i*Recall_i}{Precision_i + Recall_i},$$

$$Precision_i = \frac{TP_i}{TP_i + FP_i}, \quad Recall_i = \frac{TP_i}{TP_i + FN_i}, \quad \kappa = \frac{p_o - p_e}{1 - p_e} \quad (9)$$

where $K$ is the total number of sleep stages, $M$ is the total number of samples, $p_o$ denotes the observed agreement between predicted and true labels, and $p_e$ represents the expected agreement by chance.

This comprehensive evaluation framework enables us to not only assess the overall performance but also investigate the model's robustness across individual sleep stages and datasets. Notably, the class-specific performance provides insight into the model's ability to distinguish challenging stages such as N1 and REM, which are often misclassified due to their transitional and ambiguous signal characteristics.

### 4.2. Results and Analysis

The performance of NeuroLingua was evaluated on both the Sleep-EDF Expanded and ISRUC-Sleep Subset 1 datasets. Table 4 summarizes overall and per-class metrics, while Figure 7 presents the normalized confusion matrices for both datasets.

NeuroLingua consistently demonstrates robust performance across both datasets, with high accuracy and a macro F1 score. The model excels in primary sleep stages (Wake, N2, N3, REM), while most errors are confined to the N1 stage, in line with common challenges in automatic sleep staging. Qualitative analysis of hypnograms further confirms the model's sequence-level fidelity. Figure 8 visually compares the ground truth and predicted hypnograms for a typical overnight recording. The NeuroLingua model demonstrates a strong ability to track the overall structure of sleep architecture, including transitions between wake, NREM, and REM stages. Most stage transitions and dominant sleep segments are accurately captured. Discrepancies predominantly occur in short, transitional N1 and REM episodes, which aligns with the confusion matrix analysis and is consistent with known challenges in sleep stage classification. This qualitative example supports the quantitative metrics and further highlights the effectiveness of the proposed model in practical, sequence-level staging scenarios.

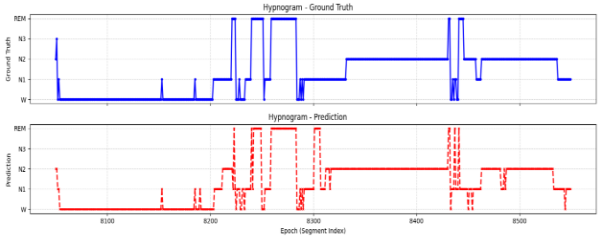

*Figure 7. Normalized Confusion Matrices (%).*

*Figure 8. Hypnogram of a portion of an overnight sleep recording from the EDF-X dataset, comparing the ground truth labels and NeuroLingua predictions.*

*Table 4. Overall and Per-Class Performance of NeuroLingua*

| Dataset | Accuracy (%) | Macro F1 | Kappa | Wake F1 | N1 F1 | N2 F1 | N3 F1 | REM F1 |
|---|---|---|---|---|---|---|---|---|
| EDF-X | 0.853 ± 0.017 | 0.800 ± 0.023 | 0.796 | 0.941 | 0.518 | 0.866 | 0.824 | 0.852 |
| ISRUC-S1 | 0.819 ± 0.041 | 0.802 ± 0.046 | 0.755 | 0.869 | 0.576 | 0.850 | 0.860 | 0.857 |

## 5. Analysis and Discussion

The purpose of this discussion is to critically interpret and contextualize the empirical findings presented in this study, with a particular emphasis on the performance, generalizability, and architectural innovations of the proposed NeuroLingua framework for sleep stage classification. This section synthesizes the main contributions and examines how they compare to current state-of-the-art approaches, offering detailed insights into the design choices and their practical implications. Specifically, the discussion covers: (i) comparative performance analysis with recent models, (ii) the contribution of each architectural component via ablation studies, (iii) the discriminative importance of individual signal modalities, and the CNN versus full NeuroLingua and (iv) a summary of the main findings across multiple datasets. By addressing both empirical results and theoretical justifications, this section aims to demonstrate the robustness, efficiency, and clinical relevance of NeuroLingua, while also identifying areas for future improvement.

### 5.1. Comparison with State-of-the-Art Methods

*Table 5. Performance comparison with other sleep staging methods on the EDF-X dataset.*

| Method | | Acc. (%) | Macro F1 | Kappa | Wake | N1 | N2 | N3 | REM | Signal |
|---|---|---|---|---|---|---|---|---|---|---|
| AttnSleep | [2] | 82.9 | 0.781 | 0.770 | 0.926 | 0.474 | 0.855 | 0.837 | 0.815 | EEG |
| MLTCN | [5] | 81.0 | 0.749 | 0.740 | 0.922 | 0.428 | 0.833 | **0.883** | 0.777 | EEG |
| CTCNet | [8] | 82.5 | 0.791 | 0.780 | 0.925 | **0.538** | 0.868 | 0.873 | 0.748 | EEG |
| EEGSNet | [11] | 83.0 | 0.773 | 0.770 | 0.932 | 0.500 | 0.842 | 0.744 | 0.835 | EEG |
| LSTM-Ladder | [14] | 84.3 | 0.770 | 0.780 | 0.920 | 0.409 | **0.885** | 0.807 | 0.823 | EEG, EOG |
| NAS | [17] | 80.0 | 0.727 | 0.720 | 0.911 | 0.392 | 0.840 | 0.810 | 0.681 | EEG |
| Sleep Transformer | [21] | 81.4 | 0.745 | 0.743 | 0.917 | 0.404 | 0.843 | 0.779 | 0.772 | EEG |
| NeuroLingua | Our | **85.3** ±1.7 | **0.800** | **0.796** | **0.941** | 0.518 | 0.866 | 0.824 | **0.852** | EEG, EOG |

*Table 6. Performance comparison with other sleep staging methods on the ISRUC-S1 dataset.*

| Method | | Acc. (%) | Macro F1 | Kappa | Wake | N1 | N2 | N3 | REM | Signal |
|---|---|---|---|---|---|---|---|---|---|---|
| MixSleepNet | [1] | **83.0** | **0.821** | **0.782** | 0.899 | **0.625** | 0.819 | **0.899** | 0.860 | EEG, EOG, ECG, EMG |
| MSF-SleepNet | [6] | 82.6 | 0.809 | 0.774 | **0.912** | 0.570 | 0.812 | 0.884 | **0.865** | EEG, EOG, ECG, EMG |
| MV-STGCN | [9] | 80.4 | 0.785 | 0.748 | 0.887 | 0.545 | 0.791 | 0.872 | 0.832 | EEG |
| MVF-SleepNet | [12] | 82.1 | 0.802 | 0.768 | 0.908 | 0.562 | 0.811 | 0.871 | 0.857 | EEG, EOG, ECG, EMG |
| U-Sleep | [15] | 77.0 | 0.770 | — | 0.890 | 0.520 | 0.790 | 0.770 | 0.880 | EEG, EOG |
| 3DSleepNet | [18] | 82.0 | 0.797 | 0.768 | 0.908 | 0.534 | 0.808 | 0.880 | 0.855 | EEG, EOG, EMG |
| SVM-ensemble | [20] | 68.4 | 0.608 | 0.583 | 0.793 | 0.242 | 0.708 | 0.808 | 0.490 | EEG |
| NeuroLingua | Our | 81.9 ±4.1 | 0.802 | 0.755 | 0.869 | 0.576 | **0.850** | 0.860 | 0.857 | EEG, EOG |

To rigorously contextualize the performance of the proposed NeuroLingua framework, we conducted comprehensive comparisons against the latest state-of-the-art (SOTA) approaches using both the Sleep-EDF Expanded (EDF-X) and ISRUC-Sleep Subset 1 (ISRUC-S1) datasets. Tables 5 and 6 provide a direct quantitative comparison with leading deep learning and hybrid models. The following discussion highlights key methodological trends and the distinctiveness of our approach.

**Sleep-EDF Expanded Dataset** – Table 5 summarizes the results of NeuroLingua and SOTA methods on the Sleep-EDF Expanded dataset. NeuroLingua achieves an overall accuracy of **85.3%**, macro F1 of **0.800**, and Cohen's kappa of **0.796**, which positions it among the strongest published methods. NeuroLingua's performance is comparable or superior to most recent SOTA methods, notably excelling in classifying challenging transitional stages (N1, N2), while delivering robust performance on underrepresented stages (REM) with a score of 0.852. Compared to methods relying solely on single-channel

EEG or limited temporal modeling (e.g., MLTCN [5], CTCNet [5], SleepTransformer [21]), NeuroLingua leverages a hierarchical temporal structure and explicit multi-modal (EEG + EOG) fusion, facilitating nuanced modeling of both local and global temporal dependencies.

**ISRUC Subgroup 1 Dataset** – Table 6 reports the comparison on the ISRUC-S1 dataset. NeuroLingua attains an overall accuracy of 81.9%, macro F1 of 0.802, and kappa of 0.755, once again positioning itself within the top tier of SOTA solutions. NeuroLingua demonstrates consistent gains over earlier single-modality models (e.g., MV-STGCN [9], SVM-ensemble [20]) and performs competitively against recent multimodal and graph-based approaches such as MixSleepNet [1], MSF-SleepNet [6], and MVF-SleepNet [12]. Notably, while MixSleepNet [1] slightly surpasses NeuroLingua in overall metrics, NeuroLingua achieves balanced per-class F1 scores, especially in N1 and REM stages, where many models suffer. Moreover, NeuroLingua achieves this strong performance using only EEG and EOG modalities, without requiring EMG or ECG signals, which some competitors do [1, 6, 12]

**Comparative Methodological Insights** – The field of automated sleep stage classification has witnessed a marked progression from early single-channel EEG models—such as SVM-ensemble approaches—to sophisticated architectures that leverage the integration of multiple physiological signals, including EEG, EOG, EMG, and ECG. In the design of NeuroLingua, a strategic focus was placed on EEG and EOG modalities, aiming to strike a balance between clinical practicality and sensor burden while maintaining high classification accuracy. By deliberately minimizing the number of required sensors, NeuroLingua offers a more feasible solution for real-world deployment without compromising performance.

Advancements in temporal modeling have been pivotal in shaping the latest generation of sleep staging models. Techniques such as hierarchical Transformers, exemplified by NeuroLingua and SleepTransformer [21], as well as RNN and GRU-based frameworks like EEGSNet [11] and 3DSleepNet [18], underscore the importance of capturing both short- and long-term temporal dependencies. NeuroLingua's two-level Transformer architecture—operating at both subwindow and segment levels—is particularly distinctive, as it is rooted in a language-inspired hierarchy and is empirically optimized in terms of context window and subwindow durations. This approach not only enhances the model's ability to capture fine-grained microstructures and broader sleep dynamics but also sets it apart in terms of design rationale and practical impact.

When considering the fusion of multi-modal information, many recent models adopt a simple feature concatenation strategy, as seen in MixSleepNet [1] and MVF-SleepNet [12]. In contrast, NeuroLingua employs a graph-based adaptive fusion mechanism, enabling dynamic modeling of interactions between modalities. This leads to a more flexible and effective integration of information, which is reflected in NeuroLingua's consistently strong performance per class, particularly in challenging sleep stages.

Interpretability has emerged as a key concern in the field, with models like MSF-SleepNet [6] and SleepTransformer [21] incorporating strategies to enhance transparency and explainability. NeuroLingua advances this trend by leveraging hierarchical structure and attention, affording interpretability at both global and fine-grained levels. This feature not only aligns with best practices but also facilitates clinical acceptance and user trust.

**Summary** – NeuroLingua stands out as a leading model on both the Sleep-EDF and ISRUC-S1 benchmarks, demonstrating remarkable performance per class and clinical applicability. Its distinctive language-inspired hierarchical temporal modeling, adaptive multi-modal fusion, and empirically tuned temporal parameters collectively differentiate it from existing state-of-the-art approaches. While specific multimodal models may report marginally higher overall accuracy, NeuroLingua's unique combination of interpretability, efficiency, and robust performance across all sleep stages positions it as a promising next-generation framework for automated sleep staging.

### 5.2. Detailed Ablation Study Results

This section analyzes the contribution of each architectural component via ablation studies.

#### 5.2.1. Replacing Transformer with LSTM

To isolate the impact of the Transformer architecture in NeuroLingua, we performed an ablation study replacing Transformers with LSTM layers while holding all other components constant. Comparative results are summarized in Table 7 and Figure 9. A detailed comparison across both the EDF-X and ISRUC-S1 datasets reveals several important insights regarding the strengths of Transformer-based models in sleep stage classification. Most notably, models utilizing Transformer architectures consistently outperform their LSTM counterparts in terms of accuracy, Macro F1 score, and detection of the diagnostically critical N1, N2, and REM stages. This performance advantage can be attributed to the Transformer's ability to model long-range and complex temporal dependencies through self-attention mechanisms, which are inherently more expressive and flexible than the sequential recurrence of LSTMs.

While LSTMs demonstrate marginally better performance in detecting wakefulness—likely due to their proficiency in modeling highly repetitive and stationary patterns—the superiority of Transformers becomes evident in the detection of N1 (a particularly ambiguous and challenging stage), as well as in N2 and REM, where subtle and nonlocal features are essential for accurate classification. The parallel, non-recurrent architecture of Transformers enables efficient processing of longer sequences, mitigates issues such as vanishing gradients, and offers greater scalability. These theoretical strengths are reflected in the consistently higher and more balanced performance of Transformer-based models across diverse datasets. From a practical standpoint, the enhanced accuracy of Transformers in clinically relevant sleep stages underscores their value for real-world sleep analysis applications. While LSTMs may be sufficient for simpler or binary classification tasks, the pronounced advantages of Transformers in handling nuanced temporal patterns and supporting robust multi-class classification clearly favor their adoption for comprehensive sleep staging.

*Table 7. Transformer vs LSTM performance comparison in the proposed NeuroLingua method.*

| Dataset | Model | Accuracy (%) | Macro F1 | Kappa | Wake | N1 | N2 | N3 | REM |
|---|---|---|---|---|---|---|---|---|---|
| ISRUC-S1 | Transformer | 81.9±4.1 | 0.802 | 0.755 | 0.869 | 0.576 | 0.850 | 0.860 | 0.857 |
|  | LSTM | 79.9±3.9 | 0.770 | 0.740 | 0.898 | 0.534 | 0.781 | 0.856 | 0.790 |
| EDF-X | Transformer | 85.3±1.7 | 0.800 | 0.796 | 0.941 | 0.518 | 0.866 | 0.824 | 0.852 |
|  | LSTM | 83.9±1.5 | 0.788 | 0.778 | 0.940 | 0.484 | 0.859 | 0.810 | 0.829 |

**Summary** – These findings provide strong evidence in support of the use of Transformer-based architectures in advanced sleep staging pipelines, particularly in scenarios where the detection of subtle stage transitions and generalizability across cohorts are crucial.

### 5.2.2. Replacing GCN with Concatenation

To quantify the impact of graph-based modality fusion, we replaced NeuroLingua's GCN fusion layer with simple feature concatenation, holding all other model components constant. This ablation was conducted on both the EDF-X and ISRUC-S1 datasets. Comparative results are summarized in Table 8 and Figure 10.

A careful evaluation of fusion strategies reveals the clear superiority of GCN-based fusion over simple feature concatenation for multi-modal sleep stage classification. Across both global performance metrics and most individual sleep stages, GCN fusion consistently yielded higher accuracy and F1 scores. This advantage was particularly evident in the detection of the N1 and REM stages, where the adaptive, graph-based approach enabled more effective extraction and integration of subtle cross-modal features—an area where concatenation methods frequently underperform. Notably, for the wake and N3 stages, simple concatenation either matched or slightly surpassed GCN fusion. This can be attributed to the highly distinctive and less ambiguous patterns characterizing these stages, which reduce the dependence on nuanced cross-modal interactions. In such cases, the static nature of concatenation is often sufficient for accurate detection. The theoretical foundation for the observed superiority of GCNs lies in their ability to dynamically model interactions between modalities. By propagating contextual information through message passing, GCNs enhance cross-modal representations, enabling the network to capture complex dependencies, particularly beneficial for ambiguous or transitional stages, such as N1 and REM. This dynamic modeling stands in sharp contrast to the limitations of static concatenation, which cannot adaptively weight or propagate information between modalities. Empirical results on both healthy and clinical cohorts further confirm the generalization and robustness of GCN-based fusion, echoing findings from the broader biomedical literature on multi-modal data integration. The improvements observed in complex stages—most notably N1, N2, and REM—are especially significant from a clinical perspective, given the critical role of these stages in the diagnosis and management of sleep disorders. In summary, the integration of GCN-based, adaptive graph fusion into advanced sleep staging frameworks offers substantial empirical and theoretical advantages over traditional concatenation. These findings strongly support the use of dynamic, graph-based fusion strategies in clinical sleep analysis pipelines, particularly for nuanced and diagnostically important tasks related to sleep stage classification.

*Table 8. GCN vs concatenation performance comparison in the proposed NeuroLingua method.*

| Dataset | Model | Accuracy (%) | Macro F1 | Kappa | Wake | N1 | N2 | N3 | REM |
|---|---|---|---|---|---|---|---|---|---|
| ISRUC-S1 | GCN | 81.9±4.1 | 0.802 | 0.755 | 0.869 | 0.576 | 0.850 | 0.860 | 0.857 |
|  | Concatenation | 79.9±4.0 | 0.772 | 0.740 | 0.897 | 0.531 | 0.784 | 0.860 | 0.787 |
| EDF-X | GCN | 85.3±1.7 | 0.800 | 0.796 | 0.941 | 0.518 | 0.866 | 0.824 | 0.852 |
|  | Concatenation | 84.4±1.9 | 0.787 | 0.783 | 0.937 | 0.489 | 0.859 | 0.815 | 0.834 |

### 5.3. Modality Importance Analysis

We evaluated the discriminative power of each physiological modality—EEG (Fpz-Cz and Pz-Oz for the EDF-X dataset, and C3-A2 and F3-A2 for the ISRUC-S1 dataset) and EOG—to assess their individual contributions to NeuroLingua's performance. Comparative results are summarized in Table 9 and Figure 11. A comprehensive analysis of channel and modality contributions reveals the clear superiority of frontal EEG electrodes—such as Fpz-Cz and F3-A2—in automated sleep stage classification. These channels consistently achieved the highest overall accuracy and macro F1 scores, highlighting their central role in sleep staging. This can be attributed to their optimal sensitivity to critical sleep phenomena, including slow-wave activity, spindles, and REM patterns, making them indispensable for both clinical and research applications. While frontal EEG remains the backbone of general sleep staging, the inclusion of EOG channels proved uniquely beneficial for REM stage detection.

Across both datasets, EOG outperformed EEG channels in accurately identifying REM periods, emphasizing its utility for capturing the rapid eye movements that define this stage. Despite these strengths, all modalities demonstrated considerable difficulty in detecting the N1 stage, with F1 scores consistently below 0.50—a reflection of the inherent ambiguity and transitional nature of N1, which continues to challenge even the most advanced models. The robustness of these findings was confirmed across both healthy and clinical populations. Frontal EEG channels provided consistently strong results in all cohorts, while EOG maintained its advantage in REM identification, underscoring the generalizability of these modality-specific strengths.

From a practical and clinical perspective, these results suggest clear recommendations. Frontal EEG electrodes, particularly Fpz-Cz, are ideally suited for minimal-channel or wearable sleep monitoring systems, thanks to their comprehensive performance across all primary sleep stages. However, for applications where accurate REM detection is critical—such as in narcolepsy diagnostics—the integration of EOG channels is essential. Furthermore, the persistent challenge of N1 classification highlights the need for continued methodological innovation, with advanced techniques such as attention-based or multi-scale architectures offering promising avenues for improvement.

These empirical observations are supported by well-established theoretical foundations: anatomically, frontal EEG electrodes are positioned to capture the full spectrum of sleep-related dynamics, from high-amplitude slow waves characteristic of N3 to spindles and K-complexes in N2, as well as the mixed-frequency patterns of REM sleep [3, 59, 60]. The unique spectral and topographical information available at these sites provides a scientific basis for their observed superiority in sleep staging.

Summary – Frontal EEG channels offer the most reliable foundation for automated sleep staging, particularly when complemented by EOG for REM detection. These findings reinforce the clinical value of multi-modal fusion and firmly establish frontal electrodes as the benchmark for both advanced research and clinical-grade sleep monitoring systems.

*Table 9. Modality Importance Analysis.*

| Dataset | Modality | Accuracy (%) | Macro F1 | Kappa | Wake | N1 | N2 | N3 | REM |
|---|---|---|---|---|---|---|---|---|---|
| EDF-X | EEG **Fpz-Cz** | **82.4**±2.7 | **0.747** | **0.754** | **0.925** | 0.385 | **0.857** | **0.801** | 0.766 |
|  | EEG Pz-Oz | 79.8±2.6 | 0.705 | 0.715 | 0.924 | 0.363 | 0.832 | 0.714 | 0.691 |
|  | EOG | 79.5±2.4 | 0.697 | 0.710 | 0.908 | 0.338 | 0.817 | 0.648 | **0.776** |
| ISRUC-S1 | EEG C3-A2 | 76.3±6.2 | 0.726 | 0.692 | **0.885** | **0.473** | 0.747 | 0.833 | 0.693 |
|  | EEG **F3-A2** | **76.4**±5.9 | 0.726 | **0.693** | 0.882 | 0.464 | **0.757** | **0.834** | 0.692 |
|  | EOG (ROC A1) | 74.2±5.1 | 0.703 | 0.664 | 0.858 | 0.399 | 0.718 | 0.826 | **0.715** |

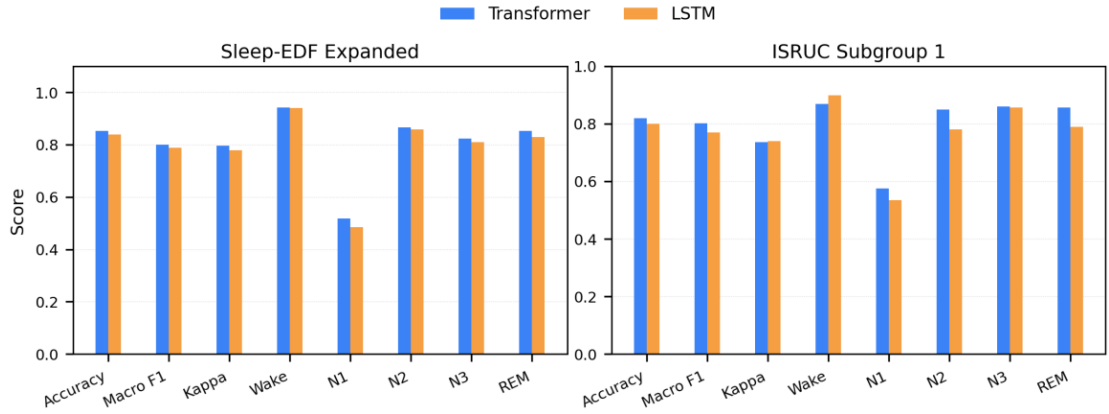

*Figure 9. Transformer vs LSTM performance comparison in proposed neurolingua method.*

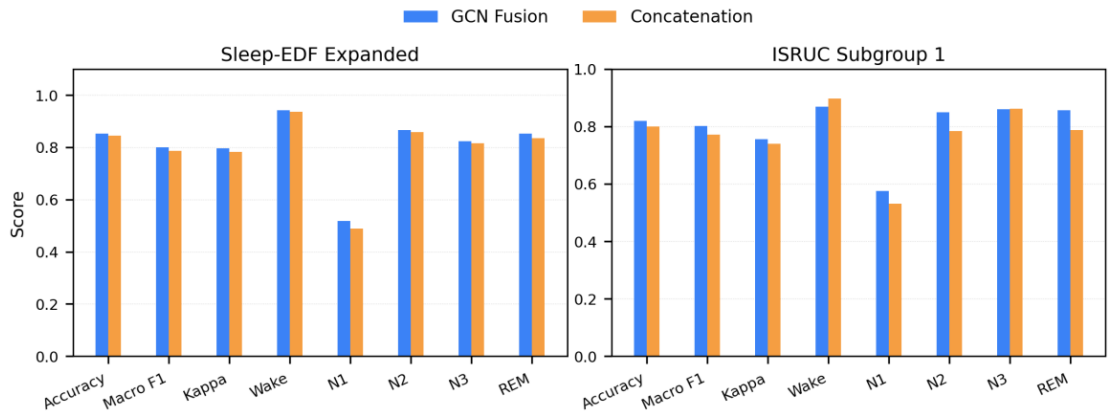

*Figure 10. GCN vs concatenation performance comparison in the proposed neurolingua method.*

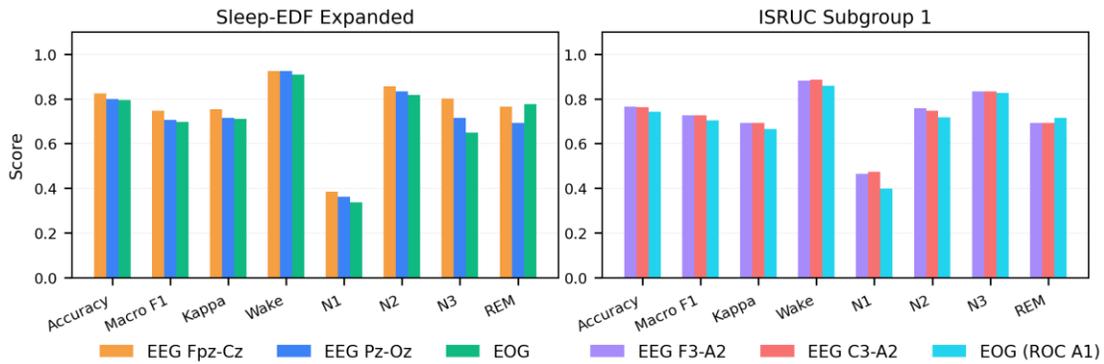

*Figure 11. Modality Importance Analysis.*

### 5.4. Only CNN vs. Full NeuroLingua

We compared the complete NeuroLingua architecture with a simplified baseline that used only the initial CNN feature extractor to quantify the impact of hierarchical and Transformer-based temporal modeling. Experiments on both EDF-X and ISRUC-S1 datasets are summarized in Table 10 and Figure 12.

The comparison between the complete NeuroLingua model and the CNN-only baseline demonstrates the critical importance of hierarchical temporal modeling for accurate sleep stage classification. Across both EDF-X and ISRUC-S1 datasets, incorporating Transformer-based intra- and inter-segment modeling leads to substantial improvements in all evaluation metrics, including accuracy, macro F1, and Cohen's kappa. Gains are particularly pronounced for the more challenging and transitional stages (N1, N2, and REM), confirming that the complete NeuroLingua architecture captures complex temporal dependencies that a simple CNN cannot. These results highlight the necessity of modeling temporal structure to achieve robust and generalizable performance in multimodal sleep staging.

*Table 10. Performance comparison with and without temporal modeling*

| Dataset | Model | Accuracy (%) | Macro F1 | Kappa | Wake | N1 | N2 | N3 | REM |
|---|---|---|---|---|---|---|---|---|---|
| EDF-X | Only CNN | 78.5 | 0.703 | 0.700 | 0.912 | 0.406 | 0.818 | 0.658 | 0.720 |
| | Full Model | **85.3** | **0.800** | **0.796** | **0.941** | **0.518** | **0.866** | **0.824** | **0.852** |
| ISRUC-S1 | Only CNN | 77.0 | 0.729 | 0.701 | **0.896** | 0.425 | 0.754 | 0.844 | 0.725 |
| | Full Model | **81.9** | **0.802** | **0.755** | 0.869 | **0.576** | **0.850** | **0.860** | **0.857** |

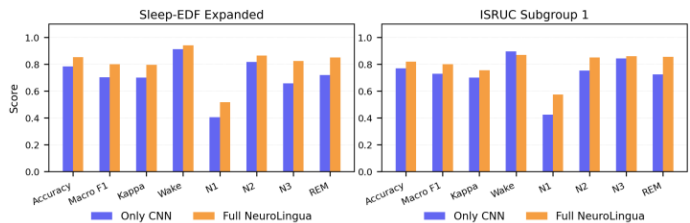

*Figure 12. Performance comparison with and without temporal modeling*

### 5.5. Summary of Main Findings

The present study establishes NeuroLingua as a powerful and generalizable framework for automated sleep stage classification, leveraging innovations in hierarchical temporal modeling and multi-modal signal integration. Beyond strong numerical performance, several key insights emerge:

1) The consistent effectiveness of NeuroLingua across both the Sleep-EDF Expanded and ISRUC Subgroup 1 datasets confirms its robust generalization capacity. The model's ability to sustain high performance with data from both healthy individuals and clinical populations demonstrates adaptability to diverse and heterogeneous sleep patterns—a critical requirement for clinical translation.

2) The model excels in accurately identifying stable sleep stages (N2, N3, and REM), which are vital for both clinical diagnosis and sleep research. This suggests that the hierarchical, language-inspired structure of NeuroLingua is well-suited for capturing salient sleep microstructures and transitions, reflecting the complex temporal dependencies inherent in sleep architecture. Significantly, the relatively lower scores for N1 reinforce longstanding challenges associated with transitional and ambiguous sleep states, underscoring the need for further methodological advances and more granular annotation standards in sleep science.

3) Finally, the cross-dataset consistency and resilience to cohort variability indicate that the core architectural principles of NeuroLingua—such as its dual-level Transformer hierarchy and adaptive graph-based fusion—successfully address several key limitations of existing approaches. This positions NeuroLingua as a promising step towards robust, interpretable, and clinically useful automated sleep staging.

### 6. Conclusion and Future Perspectives

This paper presents NeuroLingua, an innovative deep learning framework that fundamentally reimagines sleep stage classification through the lens of natural language processing. By explicitly modeling the hierarchical organization of sleep signals—drawing analogies between subwindows, segments, and extended sequences with linguistic constructs such as tokens, sentences, and paragraphs—NeuroLingua effectively captures both local and global temporal dependencies, reflecting the layered structure of natural language comprehension. This perspective, combined with hierarchical feature extraction, adaptive graph-based fusion, and advanced Transformer-based temporal modeling, establishes a new and context-aware paradigm for automated sleep analysis. Comprehensive evaluation across both healthy and clinical populations demonstrates that NeuroLingua achieves robust, balanced, and competitive performance, with consistent generalization across diverse datasets and signal modalities. While not universally superior to all state-of-the-art methods, NeuroLingua performs on par with leading approaches, and ablation studies confirm that its language-inspired hierarchical modeling and multimodal fusion provide clear benefits, particularly in classifying challenging and transitional sleep stages.

Beyond advancing the state of the art in sleep staging, NeuroLingua lays a solid foundation for future research at the intersection of language modeling and physiological signal processing. Notably, the architecture offers distinct advantages for both interpretability and explainability. In terms of interpretability, NeuroLingua's use of hierarchical attention mechanisms allows for the direct identification of which subwindows, segments, or modalities contribute most to each classification decision. This makes it possible to highlight and quantify the importance of specific temporal events or signal patterns, offering insights into how the model processes physiological data. Regarding explainability, the modular design—comprising explicit tokenization, dual-level Transformers, and graph-based multimodal fusion—enables researchers and clinicians to dissect the model's internal reasoning and understand the pathways by which predictions are formed. This structural transparency supports the generation of meaningful explanations for individual decisions, helping to uncover the rationale behind complex outputs. This combination of interpretability and explainability positions NeuroLingua as a promising foundation for reliable, transparent, and clinically meaningful automated sleep staging. As a flexible and extensible platform, it can further accommodate emerging research in areas such as causal inference and transfer learning, paving the way for enhanced clinical integration and next-generation advances in physiological signal analysis.